\documentclass[runningheads]{llncs}

\usepackage{tikz}
\usepackage{pgfplots}

\usepackage{amsmath}
\usepackage{amsfonts}
\usepackage{xcolor}
\usepackage{pgfplots}
\usepackage{subcaption}
\pgfplotsset{compat=1.17} 
\usepackage{subcaption}    

\usepackage{booktabs}     
\usepackage{graphicx}     

\usepackage{booktabs}   
\usepackage{multirow}   
\usepackage{graphicx}   
\usepackage{graphicx}    
\usepackage{subcaption}  

\usepackage{amsmath}
\usepackage{graphicx}
\usepackage{amsmath}   
\usepackage{amsfonts}  
\usepackage{amsmath}        
\usepackage{amsfonts}       
\usepackage{xcolor}         
\usepackage{pgfplots}       
\usepackage{subcaption} 
\usepackage[numbers,sort&compress]{natbib}
\definecolor{basecolor}{RGB}{76,187,239} 
\definecolor{highlightcolor}{RGB}{230,126,34} 
\definecolor{gridcolor}{RGB}{220,220,220} 

\usepackage[T1]{fontenc}
\usepackage{graphicx}
\usepackage{booktabs}

\usepackage[misc]{ifsym}

\usepackage{mwe}

\begin{document}

\title{EGAD: Entropy-Guided Adaptive Distillation for Token-Level Knowledge Transfer}

\titlerunning{EGAD}
\author{
Hao Zhang\inst{1,2,3} \and
Zhibin Zhang\inst{2} \and
Guangxin Wu\inst{2,3} \and
Wanyi Ning\inst{2} \and
Jiafeng Guo\inst{2} \and
Xueqi Cheng\inst{2}
}

\institute{
School of Advanced Interdisciplinary Sciences, University of Chinese Academy of Sciences \and
State Key Laboratory of AI Safety, Institute of Computing Technology, Chinese Academy of Sciences \and
University of Chinese Academy of Sciences
}

\maketitle              

\begin{abstract}
Large language models (LLMs) have achieved remarkable performance across diverse domains, yet their enormous computational and memory requirements hinder deployment in resource-constrained environments. Knowledge distillation offers a promising solution by transferring knowledge from a large teacher model to a smaller student model. However, existing distillation methods typically treat all tokens equally, ignoring the fact that different tokens contribute unequally to model decisions. This can lead to inefficient knowledge transfer and reduced learning effectiveness. To address this limitation, we propose an entropy-based adaptive distillation strategy that dynamically adjusts the training process at the token level. Our method leverages the teacher's output entropy to guide three aspects of distillation. Specifically, we introduce a token-level curriculum by dynamically shifting focus from low- to high-entropy tokens during training. We further adjust the distillation temperature based on token entropy to better capture teacher confidence patterns. Moreover, we employ a dual-branch architecture for efficient logits-only distillation on easy tokens and deeper feature-based distillation on difficult tokens. Extensive experiments validate the soundness and effectiveness of our method.

\end{abstract}

\section{Introduction}
In recent years, large language models (LLMs) \cite{yang2024qwen2} have achieved remarkable progress across diverse domains, largely due to increasing model size and training data. Larger models improve both generation quality and task generalization, but their high computational and memory demands limit deployment in resource-constrained settings. To mitigate this, model compression techniques such as quantization \cite{cai2023gptq}, pruning \cite{ashkboos2024slicegpt,wu2026iterative,zhang2026mi}, and knowledge distillation (KD)  have been proposed. Quantization and pruning reduce costs by lowering weight precision or removing redundant parameters, while KD transfers knowledge from a large teacher model to a smaller student, effectively balancing performance and efficiency. Its flexibility and effectiveness have made KD a widely adopted compression approach.
\begin{figure}[t]
\centering
\includegraphics[width=0.95\linewidth]{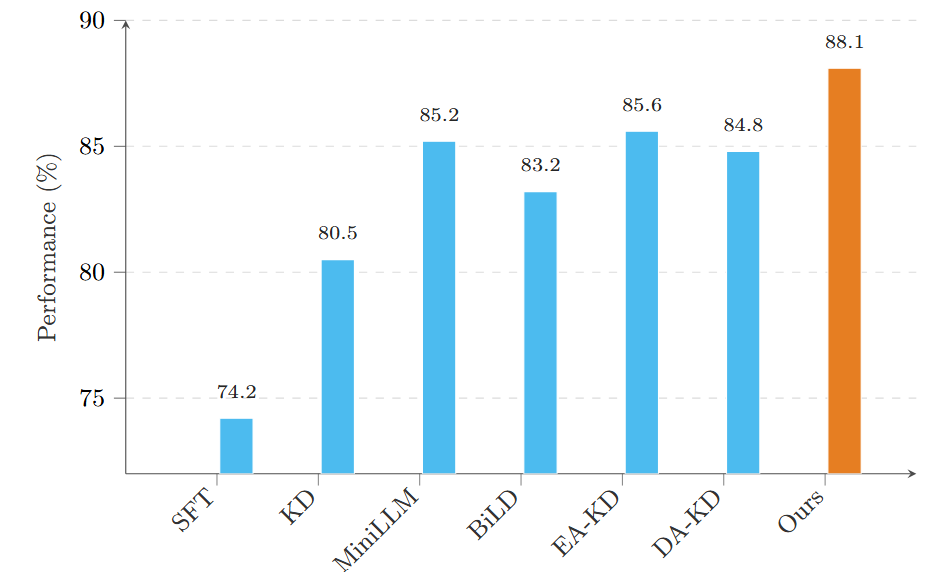}
\caption{Performance comparison of different methods on the SST-2 benchmark.}
\label{fig:distill_sst2}
\end{figure}

Although existing distillation methods can preserve student model performance to some extent, most approaches still perform uniform output matching and gradient optimization at the sequence level, without explicitly differentiating the contributions of individual tokens. Empirical studies \cite{wang2025beyond,jung2025todi} have shown that large language models exhibit a highly uneven information distribution at the token level: some tokens correspond to regions of model uncertainty and have a greater impact on the final generated outputs and downstream performance, whereas other tokens mainly involve patterned or deterministic predictions, contributing relatively little training signal. This phenomenon of information concentration has been observed across various models and data settings, often following a long-tail or “80/20” distribution. In this context, treating all tokens equally during distillation may fail to adequately transfer knowledge for high-information tokens and may even introduce redundant training signals, thereby limiting the student model’s learning efficiency and performance improvements. Therefore, modeling and leveraging token-level importance differences during distillation remains a problem that current methods have yet to systematically address.

To address the limitations caused by the indiscriminate treatment of tokens in knowledge distillation, we propose an entropy-based adaptive distillation strategy. By dynamically adjusting the training process, it enables differential learning for individual tokens. The core intuition is that the teacher model's output entropy provides a direct measure of each token's learning difficulty: low entropy corresponds to confident predictions from the teacher, which are easier for the student to grasp quickly, whereas high entropy indicates uncertainty, requiring more precise guidance. Based on this observation, we optimize the distillation process along three key dimensions: (1) We dynamically adjust the KL divergence weight for each token based on the teacher's output entropy. During the early stage of training, distillation emphasizes low-entropy tokens, enabling the student model to quickly grasp fundamental knowledge. As training progresses, attention gradually shifts to high-entropy tokens, guiding the student to tackle more challenging tokens, thereby naturally forming an entropy-based curriculum learning strategy. (2) We further associate the distillation temperature with output entropy: low-entropy tokens use a lower temperature to preserve the teacher's sharp and confident distribution, while high-entropy tokens use a higher temperature to amplify uncertainty, helping the student capture subtle probability differences and naturally learn the teacher's "certainty patterns."  (3) For tokens with different entropy levels, we adopt a differentiated distillation strategy. Low-entropy tokens undergo only logits distillation to save computational resources, whereas high-entropy tokens additionally distill intermediate layer features and attention distributions,. This is realized via a dual-branch architecture: a fast path handles low-entropy tokens, while a deep distillation path focuses on high-entropy tokens. Extensive experiments validate the soundness and effectiveness of our method. Figure~\ref{fig:distill_sst2} presents a performance comparison of different distillation methods, where our approach achieves the best results. 

Our contributions are summarized as follows:
\begin{itemize}
    \item \textbf{Entropy-aware token-level curriculum learning.} 
    We propose an entropy-guided distillation paradigm that operates at the token level, dynamically reweighting distillation objectives according to the teacher’s predictive uncertainty. This mechanism naturally induces a curriculum learning process that progressively shifts the student’s focus from confidently predicted tokens to more ambiguous and challenging ones.

    \item \textbf{Adaptive entropy-conditioned temperature scaling.} 
    We introduce an entropy-dependent temperature scheduling strategy that modulates the sharpness of the distillation targets on a per-token basis, enabling the student to faithfully inherit the teacher’s confidence structure while better modeling uncertainty in high-entropy regions.

    \item \textbf{Differentiated multi-path distillation architecture.} 
    We design a dual-branch distillation framework that selectively applies lightweight logits-based distillation to low-entropy tokens and richer feature- and attention-level supervision to high-entropy tokens.
\end{itemize}

\section{Related Work}

\subsection{Large Language Models}
Recent progress in large language models (LLMs) and vision-language foundation models has advanced model security, efficient tuning, cross-modal learning, and domain-specific intelligent applications. In terms of model security and human-machine interaction, existing studies have uncovered LLM jailbreaking vulnerabilities in benign generation behaviors and improved intent understanding for ambiguous prompts under human-machine collaborative paradigms \cite{wu2025sugar,he2025enhancing}. To enhance training efficiency and generalization performance, effective instruction tuning strategies and adaptive fine-tuning metrics have been developed to stabilize and optimize LLM training \cite{zhang2026guiding,yu2026probability}. Extensive efforts have also promoted multimodal and graph-crossed learning capabilities of foundation models, covering asymmetric expert specialization for vision-language models, text-driven story visualization, robust graph-text alignment, bilingual graph grammar modeling, and multi-agent geolocalization frameworks \cite{zhang2025asymoe,zu2026end,zhang2025can,zheng2025g2rammar,zheng2025graphgeo}. In practical downstream domains, foundation models have achieved superior performance in medical intelligence, legal analysis, and biological prediction, including medical image augmentation, pathological prognosis analysis, bone density estimation, psychiatric classification, anti-hallucination medical visual question answering, legal judgment inference, and interpretable RNA modification prediction \cite{qi2025mediaug,luo2025pathohr,qi2025medconv,cong2025hierarchical,jiang2026multi,kang2026multimodal,wang2026evormd}. Furthermore, a series of robust learning techniques support reliable model deployment in diverse scenarios, encompassing protected code representation learning, revocable multimodal sentiment analysis, multi-scale ship detection, open-vocabulary object detection, incremental remote sensing segmentation, and self-supervised feature prototyping \cite{mo2026shieldedcode,fu2026missing,hu2026p2r,wang2026deco,wu2026protoflow,zhou2026hot}.

\subsection{Knowledge Distillation}
Knowledge distillation (KD) was originally proposed as a model compression technique to transfer knowledge from a large teacher model to a smaller student model \cite{hinton2015distilling}. By matching the teacher’s softened output distributions, the student captures richer inter-class similarity information than with hard labels alone, leading to improved generalization and efficiency \cite{ba2014deep}. In the language modeling domain, the DeepSeek team introduced a two-stage distillation framework that transfers the capabilities of the 671B-parameter DeepSeek-R1 model to a 70B-scale LLaMA base model, demonstrating that carefully designed pipelines can effectively compress extremely large models while preserving substantial performance \cite{guo2025deepseek}. Moreover, the Active Mutual Distillation (AMD) module, first proposed for vision tasks, has been extended to multimodal scenarios by selectively extracting task-relevant knowledge from reliable modalities to guide less reliable ones, thereby improving robustness and overall performance \cite{wanyan2023active}. Together, these advances illustrate the versatility of KD across architectures and modalities in enhancing efficiency and generalization.

\subsection{Entropy in KD}
Several studies have incorporated entropy into knowledge distillation to better regulate knowledge transfer from teacher to student models. For example, a work \cite{cheng2020explaining} introduces entropy-based metrics to quantify knowledge retention during distillation, offering a principled measure of uncertainty preservation when compressing a teacher into a student. Building on this, AKD \cite{kwon2020adaptive} proposes adaptive weighting in multi-teacher settings, assigning greater importance to lower-entropy (more confident) predictions; however, this strategy may become suboptimal in single-teacher scenarios by overemphasizing certain outputs while overlooking informative high-entropy signals. DynamicKD \cite{zhu2024dynamickd} improves single-stage distillation via entropy correction, adopting a mechanism related to CTKD and label smoothing (LS) to adjust logit contributions based on uncertainty, though its logit-level formulation limits flexibility. TTM \cite{zheng2024knowledge} removes the student temperature during training, implicitly introducing Rényi entropy regularization to better align uncertainty structures, while RTTM further prioritizes high-uncertainty samples to enhance learning on difficult instances. More recently, EA-KD \cite{wei2025multi} dynamically weights predictions by considering both teacher and student entropy, enabling a more balanced and nuanced transfer process. We also include comparisons with these entropy-based distillation methods in our experiments.

\section{Method}
\begin{figure*}
    \centering
    \includegraphics[width=0.98\linewidth]{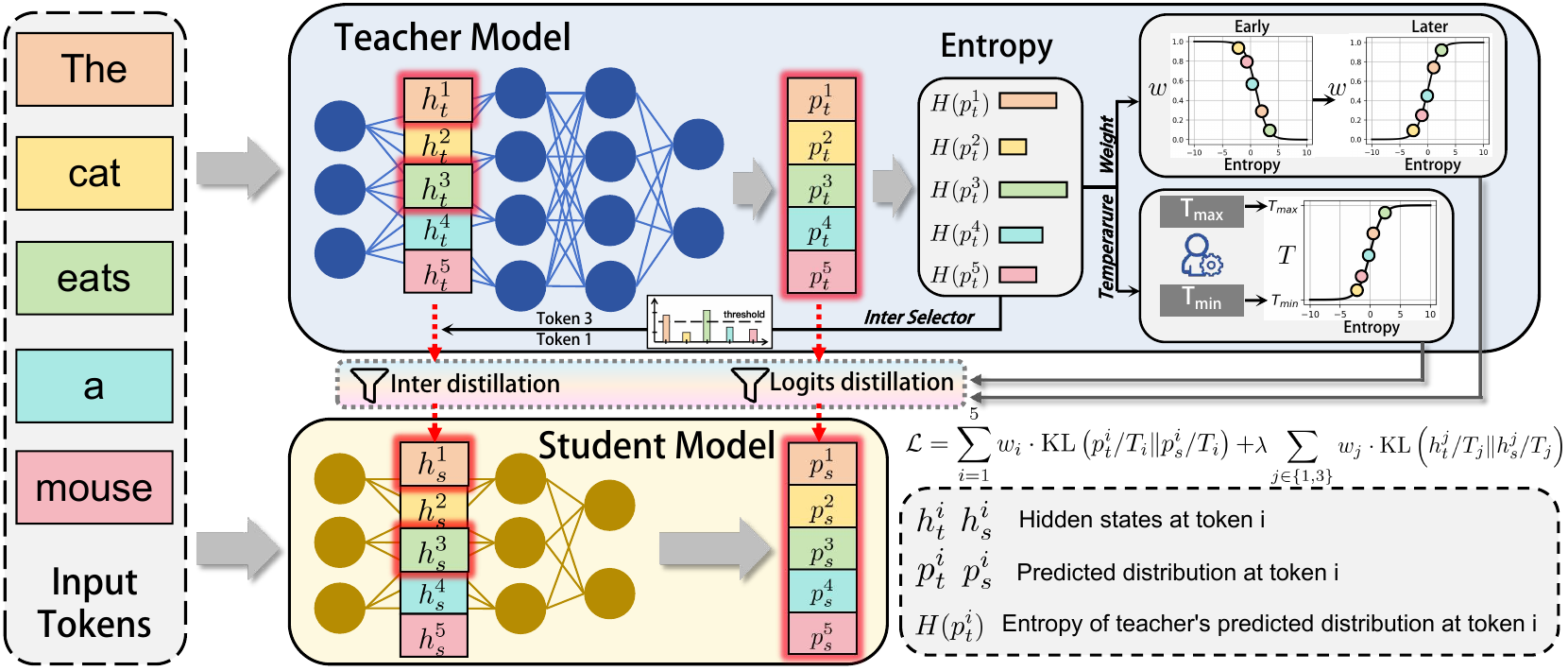}
    \caption{{{Overview of EGAD.} Given an input sequence, the teacher model produces token-level hidden states and predictive distributions, whose entropy is used as a token-wise uncertainty signal. This entropy jointly controls three adaptive components: (1) an entropy-guided curriculum that gradually shifts training emphasis from low-entropy tokens in the early stage to high-entropy tokens in the later stage, (2) a token-specific temperature schedule that adjusts the softness of distillation targets between $T_{\min}$ and $T_{\max}$, and (3) a differentiated distillation path that applies lightweight logits distillation to easy tokens and additional intermediate-representation supervision to difficult tokens.}}
    \label{fig:placeholder}
\end{figure*}

In this section, we present EGAD, an entropy-guided adaptive distillation framework for token-level knowledge transfer. As illustrated in Figure \ref{fig:placeholder}, EGAD uses teacher predictive entropy to model token-wise difficulty and adapt the curriculum, temperature, and distillation path accordingly. This enables the student to learn easy tokens first and progressively focus on harder ones, improving both efficiency and transfer quality. \textbf{We provide the theoretical analysis of our design in Appendix~\ref{sec:advanced_theory}.}
\subsection{Entropy Guided Curriculum Learning}
Knowledge distillation often treats all tokens equally, overlooking that the teacher’s predictive confidence can vary substantially across tokens. The entropy of the teacher’s output distribution provides an effective token-level proxy for learning difficulty: low-entropy tokens correspond to high-confidence, easier-to-mimic predictions, whereas high-entropy tokens reflect uncertainty and are typically harder to learn. Motivated by this observation, we propose \emph{entropy guided curriculum distillation}, which dynamically adjusts token-level distillation weights based on entropy, thereby implementing a curriculum that naturally progresses from easy to hard.

Let the teacher model’s predicted probability distribution over the vocabulary at token $i$ be $p_t^i$, and the student’s predicted distribution at the same token be $p_s^i$, where $i$ indexes tokens in the sequence. The standard knowledge distillation loss $\mathcal{L}_{\text{KD}}$ is defined as:
\begin{equation}
\mathcal{L}_{\text{KD}} = \sum_{i} \text{KL}(p_t^i \| p_s^i)
\end{equation}
where $\text{KL}(\cdot \|\cdot)$ denotes the Kullback--Leibler divergence between two probability distributions. To quantify token-wise uncertainty, we compute the entropy of the teacher’s distribution at token $i$:
\begin{equation}
H(p_t^i) = - \sum_{v \in \mathcal{V}} p_t^i(v) \log p_t^i(v)
\end{equation}
where $\mathcal{V}$ is the vocabulary, $v$ indexes a vocabulary token, and $H(p_t^i)$ measures the teacher’s uncertainty at token $i$. Higher entropy indicates a more uncertain, difficult token, while lower entropy indicates a more confident, easier token. Based on this, we define the \emph{entropy-weighted distillation loss} $\mathcal{L}_{\text{EWD}}$:
\begin{equation}
\mathcal{L}_{\text{EWD}} = \sum_i w_i \cdot \text{KL}(p_t^i \| p_s^i)
\end{equation}
where $w_i$ is a dynamic weight controlling the contribution of token $i$, computed from its entropy and the current training stage. Specifically, we schedule $w_i$ with training progress $t$ as:
\begin{equation}
w_i(t) =
\begin{cases}
\sigma(1 - H(p_t^i)), & t < t_0 \\
\sigma(H(p_t^i)), & t \ge t_0
\end{cases}
\end{equation}
where $t$ is the current training step, $t_0$ is the curriculum transition point, and $\sigma(\cdot)$ is the sigmoid function. During the early stage ($t < t_0$), low-entropy tokens are emphasized, enabling the student to quickly absorb the teacher’s high-confidence knowledge. As training proceeds ($t \ge t_0$), high-entropy tokens receive larger weights, encouraging the student to learn more complex and uncertain patterns, thus realizing an “easy-to-hard” curriculum. Notably, this entropy-guided weighting yields a dynamic learning trajectory without requiring external sample ranking or handcrafted difficulty estimation.

\subsection{Entropy Adaptive Temperature}
Standard distillation typically uses a fixed temperature $T$ to soften the teacher’s output distribution; however, this uniform strategy can be suboptimal when token difficulty varies. For low-entropy tokens, the teacher already produces sharp and confident distributions, and an overly large temperature may unnecessarily blur informative signals. In contrast, high-entropy tokens correspond to uncertain predictions, where a higher temperature can highlight subtle probability differences and help the student better capture uncertainty patterns. To this end, we introduce an \emph{entropy adaptive temperature} mechanism that adjusts the distillation temperature at the token level according to the teacher’s output entropy.

We define the temperature for token $i$ as a monotonic function of $H(p_t^i)$:
\begin{equation}
T_i = T_{\min} + (T_{\max} - T_{\min}) \cdot \sigma(H(p_t^i))
\end{equation}
where $T_{\min}$ and $T_{\max}$ are hyperparameters specifying the minimum and maximum temperatures. Low-entropy tokens are assigned temperatures close to $T_{\min}$ to preserve the sharpness of confident predictions, whereas high-entropy tokens are assigned temperatures closer to $T_{\max}$ to better expose uncertainty structure and promote nuanced imitation. With the token-specific temperature $T_i$, we compute the distillation divergence as:
\begin{equation}
\text{KL}(p_t^i/T_i \;\|\; p_s^i/T_i)
\end{equation}
where dividing by $T_i$ indicates that the logits of both teacher and student are softened using the same token-wise temperature. This token-adaptive formulation allows the student to modulate its learning signal according to inherent token difficulty, capturing both confident and uncertain knowledge more effectively.

\subsection{Differentiated Distillation Paths}
We dynamically adjust the distillation depth at the token level using the teacher’s output entropy as a measure of prediction uncertainty. The underlying intuition is that tokens for which the teacher produces a sharp, confident distribution (low entropy) are comparatively easy, and can be effectively learned through a lightweight objective that only matches output logits. In contrast, tokens associated with a flatter distribution (high entropy) reflect ambiguity or higher complexity, where logits supervision alone may be insufficient. For these challenging tokens, we apply a deeper distillation route that supplements logits distillation with additional alignment of intermediate representations and attention distributions, providing richer guidance on how the teacher internally processes and routes contextual information. This token-adaptive strategy avoids unnecessary overhead on easy positions while strengthening supervision where the student most needs it, enabling more efficient computation allocation and better use of model capacity.

Formally, let $\mathcal{L}{\text{logits}}^i$ denote the logits distillation loss for token $i$, $\mathcal{L}{\text{feat}}^i$ the intermediate feature distillation loss, and $\mathcal{L}{\text{attn}}^i$ the attention distillation loss. The overall distillation loss for token $i$ is:
\begin{equation}
\label{7}
\scalebox{0.78}{$
\mathcal{L}i =
\begin{cases}
\mathcal{L}{\text{logits}}^i, & H(p_t^i) < H{\text{th}} \\
\mathcal{L}{\text{logits}}^i + \lambda \left( \mathcal{L}{\text{feat}}^i + \mathcal{L}{\text{attn}}^i \right), & H(p_t^i) \ge H{\text{th}}
\end{cases}
$}
\end{equation}
where $H_{\text{th}}$ is the entropy threshold separating low- and high-entropy tokens, and $\lambda$ controls the contribution of deeper distillation terms. The details of feature and attention distillation are provided in the implementation details section.

\section{Experiments}
\subsection{Experimental Setup}
We formulate instruction following~\cite{ouyang2022training} as a conditional text generation problem, in which a model produces responses conditioned on input instructions. A large model is fine tuned on a dataset consisting of instruction and response pairs and serves as the teacher. We compare different knowledge distillation methods by evaluating the instruction following performance of the student model.
\paragraph{Datasets and Metrics.}
We construct the training dataset from databricks-dolly15K \cite{DatabricksBlog2023DollyV2}, a collection of 15K human-authored instruction-response pairs. We filter out samples that exceed the model’s context window before randomly splitting 1K samples for validation, 0.5K for testing, and retaining approximately 12.5K examples for training. For model evaluation, we assess the trained models on five instruction-following datasets: DollyEval (the 500-sample test subset split from databricks-dolly15K) \cite{DatabricksBlog2023DollyV2}, SelfInst \cite{wang2023self}, VicunaEval \cite{chiang2023vicuna}, S-NI \cite{asai2024buffet}, and UnNI. UnNI consists of 10K samples randomly sampled from the core set of UNNATURALINSTRUCTIONS \cite{honovich2023unnatural}.  To assess the quality of model-generated responses, we employ two metrics. The first is Rouge-L, which captures sentence-level structural similarity through longest common subsequence statistics and is well-suited for large-scale instruction-following evaluation. The second is GPT-4 feedback. For this metric, GPT-4 compares model outputs with ground-truth answers and assigns scores ranging from 1 to 10 across four dimensions: helpfulness, relevance, accuracy, and detail richness. We then report the ratio of the model’s total score to that of the ground truth. This GPT-4 feedback metric is exclusively applied to DollyEval, SelfInst, and VicunaEval.

\begin{table*}[t]
\centering
\caption{Performance comparison of different distillation methods across models and datasets.}
\label{tab:grouped_results}
\resizebox{\textwidth}{!}{%
\begin{tabular}{ll l cc cc cc c c}
\toprule
\multirow{2}{*}{Model} &
\multirow{2}{*}{Params} &
\multirow{2}{*}{Method} &
\multicolumn{2}{c}{DollyEval} &
\multicolumn{2}{c}{SelfInst} &
\multicolumn{2}{c}{Vicuna} &
\multirow{2}{*}{S-NI} &
\multirow{2}{*}{UnNI} \\
\cmidrule(lr){4-5}\cmidrule(lr){6-7}\cmidrule(lr){8-9}
& & &
Rouge-L & GPT-4 &
Rouge-L & GPT-4 &
Rouge-L & GPT-4 &
& \\
\midrule

\multirow{9}{*}{GPT-2} & 1.5B & Teacher
& 27.1 & 56.7 & 14.5 & 41.3 & 16.7 & 47.1 & 27.1 & 30.7 \\
\cmidrule(lr){2-11}
& \multirow{8}{*}{760M} & SFT
& 25.1 & 48.9 & 11.6 & 36.2 & 15.8 & 42.2 & 21.6 & 26.3 \\
& & KD
& 25.7 & 51.6 & 12.2 & 40.5 & 16.2 & 43.5 & 24.3 & 30.2 \\
& & SeqKD
& 25.4 & 50.4 & 13.4 & 41.2 & 15.4 & 42.7 & 25.1 & 31.7 \\
& & MINILLM
& 26.2 & 51.8 & 15.7 & 42.6 & 17.1 & 44.6 & 28.4 & 36.1 \\
& & BiLD
& 25.3 & 49.3 & 12.5 & 38.4 & 15.9 & 41.1 & 22.4 & 27.9 \\
& & EA-KD
& 25.6 & 50.1 & 13.1 & 40.9 & 15.7 & 42.9 & 27.0 & 30.5 \\
& & DA-KD
& 26.1 & 51.2 & 14.3 & 41.8 & 16.5 & 43.6 & 27.6 & 31.1 \\
& & Ours
& \textbf{27.0} & \textbf{52.8} & \textbf{16.8} & \textbf{44.7}
& \textbf{17.9} & \textbf{46.8} & \textbf{29.9} & \textbf{37.2} \\
\midrule

\multirow{9}{*}{OPT} & 13B & Teacher
& 28.8 & 69.5 & 17.9 & 55.1 & 16.9 & 56.5 & 30.9 & 34.8 \\
\cmidrule(lr){2-11}
& \multirow{8}{*}{2.7B} & SFT
& 26.8 & 53.2 & 14.1 & 37.5 & 16.1 & 45.2 & 23.5 & 30.4 \\
& & KD
& 25.6 & 59.6 & 13.6 & 44.9 & 16.3 & 51.2 & 25.7 & 30.6 \\
& & SeqKD
& 27.3 & 58.4 & 13.8 & 40.2 & 16.2 & 44.2 & 25.1 & 31.4 \\
& & MINILLM
& 27.4 & 61.7 & 15.9 & 49.3 & 17.9 & 54.3 & 32.2 & 33.8 \\
& & BiLD
& 27.0 & 56.7 & 14.5 & 43.2 & 16.5 & 52.3 & 29.5 & 30.8 \\
& & EA-KD
& 26.8 & 53.3 & 14.8 & 44.1 & 16.9 & 52.8 & 29.8 & 32.5 \\
& & DA-KD
& 27.2 & 58.2 & 15.6 & 48.9 & 17.6 & 54.1 & 31.0 & 33.5 \\
& & Ours
& \textbf{28.2} & \textbf{62.2} & \textbf{16.9} & \textbf{52.3}
& \textbf{18.6} & \textbf{55.8} & \textbf{33.1} & \textbf{35.2} \\
\midrule

\multirow{9}{*}{LLaMA3} & 13B & Teacher
& 30.5 & 79.0 & 23.5 & 74.0 & 19.8 & 65.4 & 36.2 & 38.6 \\
\cmidrule(lr){2-11}
& \multirow{8}{*}{8B} & SFT
& 25.1 & 71.7 & 21.5 & 68.4 & 16.9 & 60.2 & 31.4 & 33.8 \\
& & KD
& 26.8 & 72.0 & 20.2 & 70.2 & 18.6 & 62.9 & 32.9 & 37.9 \\
& & SeqKD
& 27.8 & 72.1 & 20.9 & 71.8 & 18.2 & 61.9 & 33.7 & 37.5 \\
& & MINILLM
& 29.1 & 76.8 & 23.2 & 73.0 & 20.6 & 63.9 & 35.5 & 40.2 \\
& & BiLD
& 27.3 & 71.8 & 22.4 & 71.5 & 18.0 & 62.2 & 34.5 & 38.2 \\
& & EA-KD
& 27.6 & 72.1 & 22.7 & 72.1 & 18.3 & 62.2 & 34.6 & 38.1 \\
& & DA-KD
& 28.7 & 73.2 & 22.5 & 71.9 & 19.6 & 63.6 & 34.8 & 37.9 \\
& & Ours
& \textbf{29.8} & \textbf{77.2} & \textbf{24.1} & \textbf{74.2}
& \textbf{21.5} & \textbf{65.4} & \textbf{36.1} & \textbf{41.1} \\

\bottomrule
\end{tabular}}
\end{table*}

\paragraph{Baselines and Models.}
We compare our method with several baseline methods. Supervised fine-tuning (SFT) updates all model parameters during adaptation to downstream tasks. Knowledge Distillation (KD) \cite{sanh2019distilbert} trains the student model using the teacher model's output distribution at each token generation step. Sequence-level knowledge distillation (SeqKD) \cite{zhou2023lima} can be viewed as a form of supervised fine-tuning in which the student model is trained on sequences generated by the teacher model. MINILLM \cite{gu2023minillm} replaces the forward KL divergence with the reverse KL divergence to prevent the student model from overestimating low probability regions of the teacher distribution. BiLD \cite{li2025bild} computes the KL divergence on pairwise differences constructed from the top k logits of both teacher and student models, aiming to filter long tail noise and leverage the internal ranking information of the logits. Entropy based Adaptive Knowledge Distillation (EA-KD) \cite{su2025ea} quantifies each sample's learning value by combining the entropy of teacher and student outputs and dynamically reweights the distillation loss to emphasize high entropy samples. Difficulty aware Knowledge Distillation (DA-KD) \cite{citation-0} dynamically adjusts the distillation dataset based on sample difficulty and introduces a bidirectional discrepancy loss to stabilize optimization and handle hard samples effectively. These baselines also include other entropy-based distillation methods. We evaluate three model configurations in our experiments. Specifically, a GPT-2 model with 1.5B parameters is distilled into a GPT-2 model with 760M parameters, an OPT model with 13B parameters is distilled into a 2.7B parameter variant, and a LLaMA3 model with 13B parameters is distilled into a 8B parameter variant.
\paragraph{Implementation Details.}
All experimental evaluations are performed with the PyTorch deep learning framework \citep{paszke2019pytorch}, in combination with the Hugging Face Transformers toolkit \citep{wolf2020transformers}. The computational tasks are run on a single NVIDIA A800 GPU with 80 GB of memory. We set the batch size to 32 and train the model for 10 epochs using the AdamW optimizer \citep{loshchilov2017decoupled}, with a learning rate of \(5 \times 10^{-6}\) and a weight decay of \(1.0 \times 10^{-2}\). We set $t_0$ to half of the total training steps, with $T_{\min}=1$ and $T_{\max}=5$, and set $\lambda=0.5$. The ratio of low-entropy to high-entropy tokens is set to 1:2, that is, the threshold $H_{\text{th}}$ is defined as the 1/3 quantile. For the intermediate-state distillation in Eq.~(\ref{7}), we distill both the input features and the attention outputs from the layer at the midpoint of the model. Consistent with previous work, we use a temperature parameter of 0 and a maximum length of 512.

\subsection{Main Results}

Table~\ref{tab:grouped_results} presents the main results of different distillation methods across three teacher–student settings (GPT-2, OPT, and LLaMA3) on five instruction-following benchmarks. Overall, our entropy-guided adaptive distillation consistently outperforms all baseline approaches under every model configuration and evaluation metric, demonstrating strong effectiveness and robustness. The proposed method achieves the best performance on DollyEval, SelfInst, Vicuna, S-NI, and UnNI, with clear improvements over standard KD and SeqKD in both Rouge-L and GPT-4 evaluation scores. These results indicate that entropy-guided token-level distillation enables the student model to generate responses that are not only closer to the reference answers in surface form but also better aligned in semantic relevance and instruction-following quality. The consistency between automatic metrics and GPT-4-based evaluations further suggests that these improvements reflect genuine gains in generation quality. Notably, in the LLaMA3 13B $\rightarrow$ 8B setting, the student model even surpasses the teacher on several datasets. One possible explanation is that the distillation process not only transfers knowledge from the teacher but also filters uncertain or noisy signals in the teacher outputs through the entropy-guided adaptive mechanism, thereby providing a form of regularization and denoising. Meanwhile, the student receives more stable and fine-grained supervision signals during training, enabling it to learn a more consistent generation strategy for instruction-following tasks. As a result, the student may achieve better generalization than the teacher on certain benchmarks. Similar phenomena have also been observed in prior distillation studies, suggesting that distillation can sometimes improve model generalization beyond that of the teacher. In addition, our approach demonstrates strong effectiveness across different model scales and architectures. In the GPT-2 setting, the 760M student nearly closes the gap with the 1.5B teacher and achieves performance comparable to the teacher on DollyEval. Similar trends are observed in larger-scale settings, including OPT 13B $\rightarrow$ 2.7B and LLaMA3 13B $\rightarrow$ 8B, indicating that the proposed entropy-guided adaptive distillation strategy scales well with model size and remains effective for both moderate and large student models.

\begin{figure}[t]
\centering

\begin{subfigure}[t]{0.49\linewidth}
\centering
\includegraphics[width=\linewidth]{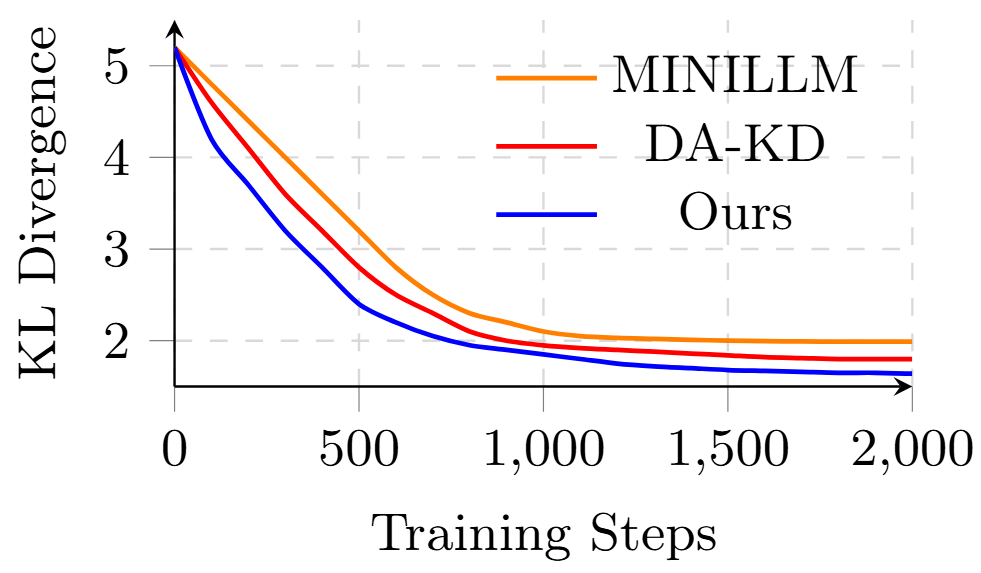}
\caption{KL divergence during training. Our method converges fastest and achieves the lowest final KL divergence, outperforming DA-KD and MINILLM.}
\label{fig:kl_loss_training_final_adjusted}
\end{subfigure}
\hfill
\begin{subfigure}[t]{0.49\linewidth}
\centering
\includegraphics[width=\linewidth]{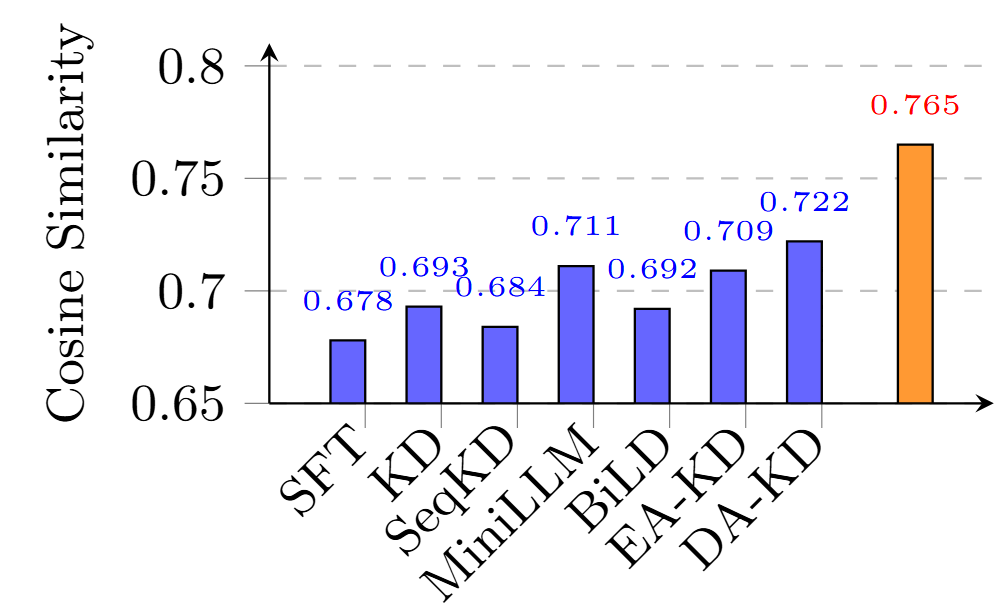}
\caption{Knowledge transfer capability is measured by the cosine similarity between the outputs of the teacher and student models on DollyEval.}
\label{fig:fidelity}
\end{subfigure}

\caption{Performance comparison among different distillation methods.}
\label{fig:distillation_comparison}
\end{figure}

Figure~\ref{fig:kl_loss_training_final_adjusted} shows the KL divergence during training for different distillation methods. As training progresses, the KL divergence gradually decreases, indicating that the student models are learning and increasingly approximating the teacher's output distribution. Among the three methods, our proposed Ours converges the fastest and achieves the lowest final KL value, demonstrating that its token-level adaptive distillation mechanism can transfer teacher knowledge more effectively. DA-KD exhibits a moderate convergence speed, while MINILLM converges more slowly and reaches the highest final KL. Overall, the trends indicate that entropy-guided adaptive strategies can improve training efficiency and better align the student model with the teacher’s distribution.

\subsection{Fidelity of Knowledge Transfer}

A key aspect of knowledge distillation is ensuring that the student model not only replicates the predictions of the teacher model but also closely captures its underlying probability distribution. To quantitatively evaluate how faithfully our student model preserves the behavior of the teacher, we perform a knowledge transfer fidelity assessment on the DollyEval dataset. Specifically, we use {Gemini}  to measure the semantic similarity between the outputs generated by our trained student model (GPT-2-760M) and those produced by the teacher model. This metric provides a fine-grained measure of alignment at the level of semantic content, going beyond simple accuracy-based comparisons. We compare our method against several representative baseline knowledge distillation approaches, including SFT, KD, SeqKD, MiniLLM, BiLD, EA-KD, and DA-KD. Each of these baselines represents a distinct strategy for transferring knowledge from teacher to student, ranging from straightforward probability matching to more advanced sequence-level or data-augmented distillation techniques. As shown in Figure~\ref{fig:fidelity}, our approach achieves a fidelity score of 76.5, outperforming all baselines by a substantial margin. Notably, the closest competitor, DA-KD, achieves a score of 72.2, highlighting that our method more effectively preserves the teacher’s semantic distribution. These results demonstrate that our approach is able to capture the nuanced patterns in the teacher’s output, which is essential for downstream tasks that rely on precise knowledge transfer.

\begin{table*}[t]
\centering
\caption{Generalization performance of different distillation methods on SST-2 and BoolQ. All student models are distilled from LLaMA-13B to LLaMA-7B using instruction-following data, without task-specific fine-tuning. Accuracy (\%) is reported.}
\label{tab:generalization_sst2_boolq}
\resizebox{0.6\textwidth}{!}{%
\begin{tabular}{lcc}
\toprule
Method & SST-2 Acc. (\%) & BoolQ Acc. (\%) \\
\midrule
SFT & 74.2 & 62.7 \\
KD & 80.5 & 63.2 \\
SeqKD & 63.4 & 64.1 \\
MiniLLM & 85.2 & 65.2 \\
BiLD & 83.2 & 64.3 \\
EA-KD & 85.6 & 63.8 \\
DA-KD & 84.8 & 65.1 \\
\textbf{Ours} & \textbf{88.1} & \textbf{66.8} \\
\midrule
Teacher & 90.9 & 71.6 \\
\bottomrule
\end{tabular}}
\end{table*}

\subsection{Generalization Performance on Unseen Classification Tasks}

Beyond instruction-following benchmarks, we further evaluate the generalization ability of distilled models on downstream classification tasks that are not directly aligned with the distillation objective. Specifically, we adopt LLaMA-13B as the teacher and distill it into a LLaMA-7B student using different distillation methods. The student models are trained solely on instruction-following data, without any task-specific fine-tuning, and are directly evaluated on SST-2 and BoolQ using classification accuracy. As shown in Table~\ref{tab:generalization_sst2_boolq}, all distillation methods outperform supervised fine-tuning, indicating effective knowledge transfer from the teacher. However, the performance varies considerably across methods. SeqKD performs poorly on SST-2, suggesting that distilling generated sequences may overfit surface-level patterns and fail to capture transferable discriminative features. In contrast, logit-based methods consistently improve generalization. Our method achieves the best performance among all student models, reaching 88.1\% accuracy on SST-2 and 66.8\% on BoolQ, and substantially narrowing the gap to the teacher. We attribute this improvement to entropy-guided token-level distillation, which enables the student to first acquire robust and transferable knowledge from low-entropy tokens and progressively learn more ambiguous decision boundaries from high-entropy tokens. These results demonstrate that entropy-guided adaptive distillation improves not only instruction-following performance but also out-of-distribution generalization on unseen downstream tasks.

\subsection{Analysis of Token Entropy Distribution}

\begin{figure}[t]
    \centering
    \begin{subfigure}[b]{0.45\linewidth}
        \centering
        \includegraphics[width=\linewidth]{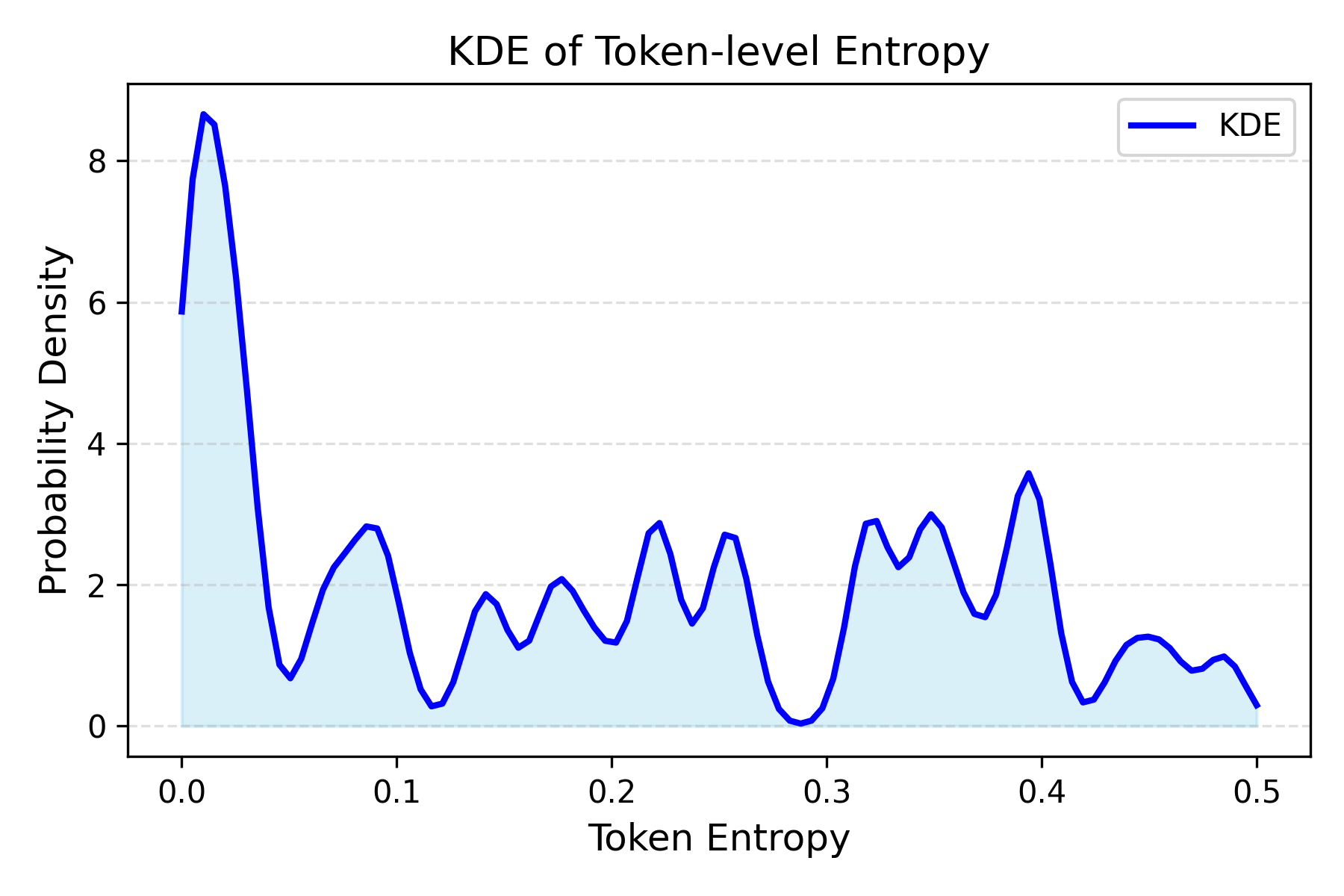}
        \caption{Sample 1 token entropy distribution.}
        \label{fig:token_entropy_sample1}
    \end{subfigure}
    \hfill
    \begin{subfigure}[b]{0.45\linewidth}
        \centering
        \includegraphics[width=\linewidth]{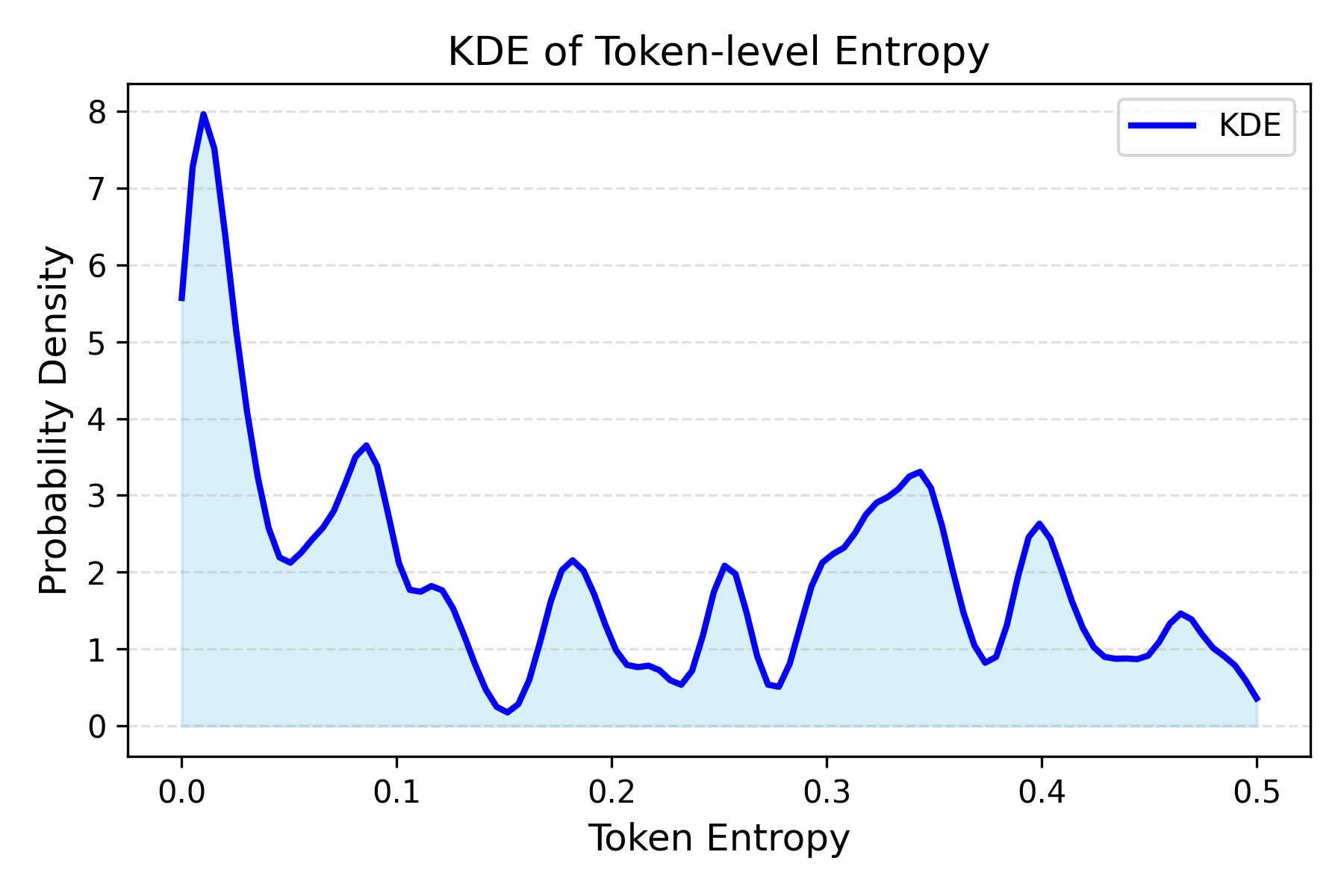}
        \caption{Sample 2 token entropy distribution.}
        \label{fig:token_entropy_sample2}
    \end{subfigure}
    \caption{Token-level entropy predicted by the teacher model for two randomly selected samples from the dataset. These distributions illustrate the variation in prediction uncertainty across different tokens.}
    \label{fig:token_entropy}
\end{figure}

To better understand the difficulty characteristics at the token level, we sampled several representative instances from the instruction-response dataset and computed the output entropy of each token predicted by the teacher model. We then visualized the distribution of token entropy using Kernel Density Estimation (KDE) to obtain a smooth approximation of its underlying probability density. The results are shown in Figure~\ref{fig:token_entropy}. It can be observed that most tokens have relatively low entropy, corresponding to predictions where the teacher model is more confident, while a small portion of tokens exhibit high entropy, indicating more difficult or ambiguous predictions. These high-entropy tokens require more precise guidance during the distillation process. This analysis further supports our proposed entropy-guided adaptive distillation strategy, which differentiates treatment for simple and difficult tokens to optimize knowledge transfer.

\subsection{Hyperparameter Analysis}

We further investigate the sensitivity of our method to key hyperparameters, including the minimum and maximum temperature $(T_{\min}, T_{\max})$, the curriculum switch point $t_0$, the high/low entropy token threshold $H_{\text{th}}$, and the feature distillation weight $\lambda$. For each hyperparameter, we vary it while keeping the others fixed and report the student model performance on DollyEval and SelfInst using Rouge-L scores. The results in Table~\ref{tab:hyperparam_analysis} show that increasing $T_{\max}$ generally improves the learning of high-entropy tokens, while excessively large values slightly degrade overall performance, and the minimum temperature $T_{\min}$ has only minor influence on low-entropy tokens. Setting the curriculum switch point $t_0$ too early reduces performance on low-entropy tokens, whereas switching too late delays the learning of difficult tokens, with the best balance achieved when $t_0$ is set to half of the training steps. A moderate entropy threshold $H_{\text{th}}$ ensures that the student focuses on challenging tokens without excessive computation, since too few high-entropy tokens underutilize the deep distillation path while too many increase cost with marginal gains. Finally, a moderate feature distillation weight $\lambda$ of 0.5 provides a good trade-off between capturing intermediate features for high-entropy tokens and maintaining training stability. Overall, these observations indicate that our method is robust to hyperparameter variations within reasonable ranges.

\begin{table}[!h]
\centering
\caption{Hyperparameter analysis of our method on DollyEval, SelfInst, and Vicuna. Each hyperparameter is varied while keeping others fixed. Best performance is highlighted in bold.}
\label{tab:hyperparam_analysis}
\resizebox{0.9\textwidth}{!}{%
\begin{tabular}{lccccc}
\toprule
Hyperparameter & Setting & DollyEval & SelfInst & Vicuna & Avg. \\
\midrule
$T_{\min}/T_{\max}$ & 1 / 3 & 27.5 & 15.8 & 17.9 & 20.4 \\
& 1 / 5 & \textbf{28.2} & \textbf{16.9} & \textbf{18.6} & \textbf{21.2} \\
& 1 / 7 & 27.9 & 16.4 & 18.3 & 20.9 \\
\midrule
$t_0$ & 1/4 & 27.8 & 16.1 & 18.2 & 20.7 \\
& 1/2 & \textbf{28.2} & \textbf{16.9} & \textbf{18.6} & \textbf{21.2} \\
& 3/4 & 27.6 & 16.3 & 18.4 & 20.8 \\
\midrule
$H_{\text{th}}$ & 1/4 high-entropy & 27.6 & 16.4 & 18.1 & 20.7 \\
& 1/3 high-entropy & \textbf{28.2} & \textbf{16.9} & \textbf{18.6} & \textbf{21.2} \\
& 1/2 high-entropy & 27.9 & 16.5 & 18.3 & 20.9 \\
\midrule
$\lambda$ & 0.25 & 27.8 & 16.3 & 18.2 & 20.8 \\
& 0.5 & \textbf{28.2} & \textbf{16.9} & \textbf{18.6} & \textbf{21.2} \\
& 1.0 & 27.7 & 16.5 & 18.4 & 20.9 \\
\bottomrule
\end{tabular}}
\end{table}

\begin{table*}[!h]
\centering
\caption{Ablation results on the OPT (13B → 2.7B) model showing the contribution of each module in our entropy-guided adaptive distillation framework. Metrics reported are Rouge-L scores. Removing any component reduces performance, confirming the effectiveness of each design choice.}
\label{tab:ablation_opt}
\resizebox{0.85\textwidth}{!}{%
\begin{tabular}{l c c c c c c}
\toprule
Variant & DollyEval & SelfInst & Vicuna & S-NI & UnNI \\
\midrule
Full model (Ours) & \textbf{28.2} & \textbf{16.9} & \textbf{18.6} & \textbf{33.1} & \textbf{35.2} \\
w/o Curriculum & 27.4 & 15.9 & 17.8 & 32.0 & 34.0 \\
w/o Adaptive Temp & 27.7 & 16.2 & 18.1 & 32.5 & 34.5 \\
w/o Diff. Paths & 27.9 & 16.5 & 18.3 & 32.7 & 34.8 \\
\bottomrule
\end{tabular}}
\end{table*}

\subsection{Ablation Study}

To assess the contribution of each component in our entropy-guided adaptive distillation framework, we conduct ablation experiments on the OPT student model (2.7B). We evaluate the impact of removing the entropy-guided curriculum, entropy-adaptive temperature, and differentiated distillation paths individually, while keeping the other components intact. Experiments are conducted on the same five instruction-following datasets used in the main results: DollyEval, SelfInst, Vicuna, S-NI, and UnNI. We use both Rouge-L and GPT-4 feedback scores as evaluation metrics. As shown in Table~\ref{tab:ablation_opt}, removing any of the three core components—entropy-guided curriculum, entropy-adaptive temperature, or differentiated distillation paths—leads to a consistent drop in performance across all datasets in terms of Rouge-L score. The curriculum contributes most notably to learning efficiency on challenging datasets such as SelfInst and Vicuna, whereas adaptive temperature and differentiated paths further enhance token-level knowledge transfer. These results confirm that our combined design effectively enables the student model to capture both easy and difficult knowledge from the teacher, improving instruction-following ability.

\section{Conclusion}
In this work, we introduce an entropy-based adaptive distillation strategy to overcome the limitation of conventional knowledge distillation that assigns uniform importance to all tokens. Using the teacher model’s output entropy as an uncertainty-aware signal, our approach modulates token-level supervision through three complementary components: an entropy-guided curriculum, an entropy-adaptive temperature schedule, and a dual-branch differentiated distillation pathway. Extensive experiments demonstrate that our method substantially strengthens knowledge transfer, allowing compact student models to more faithfully inherit the teacher’s critical capabilities and representations.

%
%
%
\bibliographystyle{splncs04}
\bibliography{paper}
%

\clearpage

\appendix

\section{Theoretical Analysis of Entropy-Guided Adaptive Distillation}
\label{sec:advanced_theory}

In this appendix, we provide a rigorous and comprehensive theoretical analysis of the proposed {Entropy-Guided Adaptive Distillation (EGAD)} framework. Our objective is to formally characterize how entropy-guided token weighting, entropy-adaptive temperature scaling, and differentiated distillation paths jointly shape (i) the token-level gradient dynamics, (ii) the stability and convergence behavior of stochastic optimization, and (iii) the overall efficiency of knowledge transfer from teacher to student. Beyond an optimization-centric view, we further analyze EGAD through an information-theoretic lens, making explicit connections between token-level entropy modulation and improved utilization of informative supervisory signals. These analyses collectively provide deeper intuition for why EGAD yields more stable training and more effective transfer under limited student capacity.

\subsection{Token-Level Gradient Dynamics and Variance Reduction}

Consider a teacher model $f_T$ and a student model $f_S$, predicting token distributions $p_t^i$ and $p_s^i$, respectively, for a given token $i$. In classical knowledge distillation, the student is trained to minimize the Kullback-Leibler (KL) divergence between the teacher and student distributions:
\begin{equation}
\mathcal{L}_{\text{KD}}^i = \sum_{v\in \mathcal{V}} p_t^i(v) \log \frac{p_t^i(v)}{p_s^i(v)},
\end{equation}
where $\mathcal{V}$ denotes the vocabulary. The gradient of this loss with respect to the student parameters $\theta$ is
\begin{equation}
\nabla_\theta \mathcal{L}_{\text{KD}}^i = - \sum_{v\in \mathcal{V}} \frac{p_t^i(v)}{p_s^i(v)} \nabla_\theta p_s^i(v).
\end{equation}

The variance of the stochastic gradient across a batch of $N$ tokens is given by
\begin{equation}
\small
\mathrm{Var}[\nabla_\theta \mathcal{L}_{\text{KD}}] = \frac{1}{N}\sum_{i=1}^N \left\|\nabla_\theta \mathcal{L}_{\text{KD}}^i - \overline{\nabla_\theta \mathcal{L}_{\text{KD}}}\right\|^2,
\end{equation}
where $\overline{\nabla_\theta \mathcal{L}_{\text{KD}}}$ denotes the mean gradient over the batch. Importantly, the magnitude and dispersion of $\nabla_\theta \mathcal{L}_{\text{KD}}^i$ can vary substantially across tokens. Empirically and theoretically, it is well-known that high-entropy tokens—where $p_t^i$ is closer to uniform and the teacher assigns comparable probability mass to many candidates—tend to induce higher gradient variance. Intuitively, when the teacher distribution is uncertain, small perturbations in the student probabilities can lead to larger relative ratios $\frac{p_t^i(v)}{p_s^i(v)}$ for multiple vocabulary items, producing noisy, competing gradient signals. Such elevated variance can destabilize early-stage optimization, slow down progress, and in practice may cause oscillatory behavior or suboptimal local trajectories, especially when the student is under-parameterized.

EGAD introduces a token-level weighting function $w_i(t)$, which depends both on the entropy of the teacher prediction $H(p_t^i)$ and the current training step $t$:
\begin{equation}
\label{eq:weighted_loss}
\mathcal{L}_{\text{EGAD}} = \sum_{i=1}^N w_i(t) \cdot \mathcal{L}_i(T_i),
\end{equation}
where $T_i$ denotes the entropy-adaptive temperature for token $i$. By explicitly modulating the contribution of each token to the total loss, EGAD reshapes the effective sampling distribution over tokens: in the early stage, the optimizer primarily receives gradients from low-entropy (high-confidence) tokens, and only later increasingly incorporates high-entropy (hard) tokens. This curriculum effect reduces gradient noise at the beginning of training, allowing the student to first establish a stable predictive backbone aligned with the teacher’s confident knowledge, which can serve as a more reliable foundation for learning ambiguous cases later.

Formally, the expected gradient norm is bounded as
\begin{equation}
\small
\mathbb{E}[\|\nabla_\theta \mathcal{L}_{\text{EGAD}}\|^2] \le \sum_{i=1}^N w_i^2(t) \, \mathbb{E}[\|\nabla_\theta \mathcal{L}_i(T_i)\|^2],
\end{equation}
indicating that an appropriate schedule of $w_i(t)$ controls the contribution of tokens whose gradients are large or highly variable. In particular, down-weighting high-entropy tokens at early steps suppresses the influence of noisy gradients, mitigating risks such as exploding updates or highly fluctuating directions. As training progresses and the student becomes better calibrated, increasing weights on high-entropy tokens gradually exposes the student to richer, more nuanced teacher information while maintaining optimization stability.

\subsection{Entropy-Adaptive Temperature as a Gradient Preconditioner}

A key novelty of EGAD lies in adapting the softmax temperature $T_i$ based on the token’s entropy. Concretely, the teacher distribution with temperature $T_i$ is
\begin{equation}
p_t^i(v; T_i) = \frac{\exp(z_v^i / T_i)}{\sum_{u\in\mathcal{V}} \exp(z_u^i / T_i)},
\end{equation}
where $z_v^i$ is the teacher logit for token $v$. The gradient of the token loss with respect to the student parameters becomes
\begin{equation}
\begin{aligned}
\nabla_\theta \mathcal{L}_i(T_i)
&= \frac{1}{T_i^2}
\sum_{v\in\mathcal{V}}
\Big(
p_s^i(v;T_i)-p_t^i(v;T_i)
\Big)\cdot \nabla_\theta z_s^i(v).
\end{aligned}
\end{equation}
This expression reveals two coupled effects of temperature. First, the distributions $p_t^i(\cdot;T_i)$ and $p_s^i(\cdot;T_i)$ become softer as $T_i$ increases, which changes the relative emphasis across vocabulary items (e.g., giving non-top candidates more influence). Second, the prefactor $\frac{1}{T_i^2}$ rescales the gradient magnitude, directly affecting step sizes along the corresponding token-induced direction.

This can be interpreted as a \textbf{diagonal preconditioning operator} applied to the gradient:
\begin{equation}
\mathbf{G}_i(T_i) = \frac{1}{T_i^2} \mathbf{J}_s^\top \mathbf{J}_s,
\end{equation}
where $\mathbf{J}_s$ is the Jacobian of the student logits with respect to parameters. From this view, temperature adaptation acts as a token-dependent conditioning mechanism that modulates both the curvature experienced by SGD and the relative sensitivity of parameters to mismatches between teacher and student. Low-entropy tokens (with small $T_i$) preserve sharp teacher signals and induce stronger gradients concentrated on the most likely targets, reinforcing confident predictions and accelerating early alignment. In contrast, high-entropy tokens (with larger $T_i$) soften the distributions, allowing gradients to reflect finer-grained differences among many plausible candidates; this encourages the student to capture the teacher’s uncertainty structure rather than overfitting to a single mode. Overall, entropy-adaptive temperature improves gradient conditioning across tokens of heterogeneous difficulty, facilitating progress along challenging directions in parameter space without destabilizing updates on easier tokens.

\subsection{Differentiated Distillation Paths and Hierarchical Feature Alignment}

EGAD also leverages a \textbf{dual-path distillation mechanism} to more effectively transfer knowledge from teacher to student, especially for high-entropy tokens. The token-level loss is defined as
\begin{equation}
\mathcal{L}_i = \mathcal{L}_{\text{logits}}^i + \lambda \big(\mathcal{L}_{\text{feat}}^i + \mathcal{L}_{\text{attn}}^i \big),
\end{equation}
where $\mathcal{L}_{\text{feat}}^i$ and $\mathcal{L}_{\text{attn}}^i$ align the student’s intermediate representations $\phi_S^l$ and attention maps $A_S^l$ with those of the teacher $\phi_T^l$ and $A_T^l$:
\begin{align}
\mathcal{L}_{\text{feat}}^i &= \sum_{l \in \mathcal{L}_d} \|\phi^l_T(x_i) - \phi^l_S(x_i)\|_2^2, \\
\mathcal{L}_{\text{attn}}^i &= \sum_{l \in \mathcal{L}_d} \|A^l_T(x_i) - A^l_S(x_i)\|_F^2.
\end{align}

This hierarchical alignment encourages the student not only to match the teacher’s output behavior (logits), but also to approximate the teacher’s internal feature geometry and attention allocation patterns. Intuitively, logits-level matching constrains the final decision boundary, whereas feature- and attention-level matching constrains intermediate computations that produce those decisions. This is particularly valuable for high-entropy tokens, where multiple alternatives are plausible and the teacher’s internal representations contain richer relational information about context, long-range dependencies, and competing hypotheses. By aligning internal structures, the student can reduce representational mismatch and improve generalization, even when its capacity limits prevent it from exactly replicating the teacher’s final-layer distribution.

The total approximation error over a batch is upper-bounded by
\begin{equation}
\begin{aligned}
\epsilon^2
&= \sum_{i=1}^N \|f_T(x_i) - f_S(x_i)\|^2 
\le \sum_{i=1}^N \Big(
\|\mathcal{L}_{\text{logits}}^i\|^2
+ \lambda^2 
\cdot
\|\mathcal{L}_{\text{feat}}^i + \mathcal{L}_{\text{attn}}^i\|^2
\Big).
\end{aligned}
\end{equation}
highlighting that deep-path alignment is particularly beneficial for high-entropy, information-rich tokens. In other words, when the teacher is uncertain, matching internal representations provides additional constraints that help the student converge toward teacher-like computations, thereby reducing the effective hypothesis space and improving the fidelity of transferred knowledge.

\subsection{Convergence Analysis under Stochastic Approximation}

Let $\theta_t$ denote the student parameters at step $t$, and assume that $\mathcal{L}_{\text{EGAD}}$ is $L$-smooth and bounded below. The standard stochastic gradient update is
\begin{equation}
\theta_{t+1} = \theta_t - \eta_t \nabla_\theta \mathcal{L}_{\text{EGAD}}(\theta_t),
\end{equation}
with learning rate $\eta_t$. Under typical stochastic approximation conditions, the expected gradient norm satisfies
\begin{equation}
\scalebox{0.8}{$
\begin{aligned}
\frac{1}{T} \sum_{t=1}^T
\mathbb{E}\!\left[
\left\|
\nabla_\theta \mathcal{L}_{\text{EGAD}}(\theta_t)
\right\|^2
\right]
&\le
\frac{2\big(\mathcal{L}_{\text{EGAD}}(\theta_0)-\mathcal{L}_{\min}\big)}
{\sum_{t=1}^T \eta_t}
+ \frac{L}{T}
\sum_{t=1}^T \eta_t \,
\mathrm{Var}\!\left[
\nabla_\theta \mathcal{L}_{\text{EGAD}}
\right].
\end{aligned}
$}
\end{equation}
where the second term is strongly influenced by the variance of the stochastic gradient. EGAD directly targets this variance term through two complementary levers: (i) curriculum-style token weighting reduces the early contribution of high-entropy tokens that would otherwise dominate gradient variance, and (ii) entropy-adaptive temperature improves conditioning and stabilizes gradient magnitudes across heterogeneous tokens. Consequently, the optimization trajectory becomes smoother and less sensitive to mini-batch composition, yielding faster practical convergence and improved stability.

Moreover, the dual-path distillation further stabilizes gradient dynamics for high-entropy tokens by providing additional supervision at intermediate layers. This added guidance can reduce the variance of gradients propagated to earlier layers (by constraining internal representations), which in turn improves asymptotic behavior and helps the student reach better minima under limited capacity. Taken together, these mechanisms suggest that EGAD can achieve both improved convergence speed and stronger final performance, consistent with empirical observations.

\subsection{Information-Theoretic Perspective}

From an information-theoretic viewpoint, the teacher’s output distribution reflects the conditional uncertainty of the next token given context, and is closely related to the conditional entropy $H(Y|X)$ of the target sequence given the input. Conventional knowledge distillation effectively assumes that all tokens contribute equally to information transfer, regardless of whether the teacher is confident or uncertain. In contrast, EGAD introduces entropy-guided weighting, which can be interpreted as reallocating learning emphasis to improve the efficiency of transferring informative signals under a finite student capacity and a finite optimization budget.

Concretely, EGAD prioritizes tokens with higher teacher entropy (at appropriate stages of training), which are more likely to encode richer relational structure (e.g., multiple plausible continuations, syntactic ambiguity, or semantic alternatives). This emphasis can be related to maximizing the mutual information between teacher and student outputs:
\begin{equation}
\small
\begin{aligned}
I(Y_{\text{teacher}}; Y_{\text{student}})
&= \sum_{i=1}^N
w_i\, H\!\left(Y_{\text{student}}^i\right)
- H\!\left(
Y_{\text{student}}^i \mid Y_{\text{teacher}}^i
\right).
\end{aligned}
\end{equation}
where $w_i$ is proportional to the entropy of the teacher prediction for token $i$. Under this formulation, increasing $w_i$ for information-rich tokens encourages the student to allocate representational capacity to match teacher behavior on tokens that carry more uncertainty and potentially more transferable structure. Simultaneously, minimizing the conditional term $H\!\left(Y_{\text{student}}^i \mid Y_{\text{teacher}}^i\right)$ corresponds to reducing student uncertainty given the teacher signal, i.e., improving fidelity of imitation. Therefore, entropy-guided weighting can be viewed as a principled mechanism to improve mutual-information efficiency: it does not merely increase training signal uniformly, but emphasizes tokens that are more valuable for transferring the teacher’s nuanced predictive distribution, ultimately resulting in more efficient and effective knowledge distillation.

\section{Intermediate Output Alignment on High-Entropy Tokens}
\label{app:2}
Beyond output-level evaluation, we further examine whether the proposed entropy-guided adaptive distillation facilitates deep representational alignment between the teacher and student models. Specifically, during the final training epoch, we focus on high-entropy tokens and measure the cosine similarity between the teacher and student representations at the middle layer, including both hidden features and attention outputs. The results are summarized in Table~\ref{tab:feature_alignment}. We observe that the cosine similarities for both intermediate features and attention distributions consistently exceed 0.9 across all evaluated settings. This indicates that, by the end of training, the student model closely matches the teacher not only at the output level but also in terms of internal representation geometry on difficult, high-uncertainty tokens. These findings provide strong evidence that the differentiated distillation paths, where feature and attention distillation are selectively applied to high-entropy tokens, are effective in transferring structural and semantic knowledge rather than merely matching logits. The high cosine similarity further explains the superior performance of our method on challenging instruction-following benchmarks and unseen downstream tasks, where accurate modeling of uncertainty and nuanced decision boundaries is crucial.

\begin{table*}[t]
\centering
\caption{Cosine similarity between teacher and student representations on high-entropy tokens at the middle layer during the final training epoch.}
\label{tab:feature_alignment}
\resizebox{0.8\textwidth}{!}{
\begin{tabular}{lcc}
\toprule
 & Feature Cosine & Attention Cosine \\
\midrule
High-Entropy Tokens & 0.93 & 0.91 \\
\bottomrule
\end{tabular}
}
\end{table*}

\end{document}